\documentclass{article}

\usepackage{microtype}
\usepackage{graphicx}
\usepackage{subfigure}
\usepackage{booktabs} 
\usepackage{diagbox}
\usepackage{bbding}

\usepackage{hyperref}

\usepackage[accepted]{icml2024}

\usepackage{amsmath}
\usepackage{amssymb}
\usepackage{mathtools}
\usepackage{amsthm}
\usepackage{multirow}

\usepackage[capitalize,noabbrev]{cleveref}

\theoremstyle{plain}
\newtheorem{theorem}{Theorem}[section]

\theoremstyle{definition}

\theoremstyle{remark}

\usepackage[textsize=tiny]{todonotes}

\icmltitlerunning{Sparse is Enough in Fine-tuning Pre-trained Large Language Models}

\begin{document}

\twocolumn[
\icmltitle{Sparse is Enough in Fine-tuning Pre-trained Large Language Models}



\icmlsetsymbol{equal}{*}

\begin{icmlauthorlist}
\icmlauthor{Weixi Song}{equal,whu}
\icmlauthor{Zuchao Li}{equal,whu,ljlab}
\icmlauthor{Lefei Zhang}{whu,ljlab}
\icmlauthor{Hai Zhao}{sjtu}
\icmlauthor{Bo Du}{whu,ljlab}

\end{icmlauthorlist}

\icmlaffiliation{whu}{National Engineering Research Center for Multimedia Software, School of Computer Science, Wuhan University, Wuhan, 430072, P. R. China}
\icmlaffiliation{ljlab}{Hubei Luojia Laboratory, Wuhan 430072, P. R. China}
\icmlaffiliation{sjtu}{Department of Computer Science and Engineering, Shanghai Jiao Tong University, Shanghai, 200240, P. R. China}

\icmlcorrespondingauthor{Zuchao Li}{zcli-charlie@whu.edu.cn}

\icmlkeywords{Parameter-Efficient Fine-Tuning, Deep Learning, Machine Learning}

\vskip 0.3in
]



\printAffiliationsAndNotice{\icmlEqualContribution} 

\begin{abstract}
With the prevalence of pre-training-fine-tuning paradigm, how to efficiently adapt the pre-trained model to the downstream tasks has been an intriguing issue. \textbf{P}arameter-\textbf{E}fficient \textbf{F}ine-\textbf{T}uning (PEFT) methods have been proposed  for low-cost adaptation. Although PEFT has demonstrated effectiveness and been widely applied, the underlying principles are still unclear. In this paper, we adopt the PAC-Bayesian generalization error bound, viewing pre-training as a shift of prior distribution which leads to a tighter bound for generalization error. We validate this shift from the perspectives of oscillations in the loss landscape and the quasi-sparsity in gradient distribution. Based on this, we propose a gradient-based sparse fine-tuning algorithm, named \textbf{S}parse \textbf{I}ncrement \textbf{F}ine-\textbf{T}uning (SIFT), and validate its effectiveness on a range of tasks including the GLUE Benchmark and Instruction-tuning. The code is accessible at \url{https://github.com/song-wx/SIFT}.
\end{abstract}

\section{Introduction}
\label{introduction}
With the prevalent pre-training-fine-tuning paradigm,  Large Language Models (LLMs) pre-trained on a internet-scale corpus can be effectively adapted to the specific downstream domains. However, as the number of parameters in large language models increases, fine-tuning all parameters is prohibitively expensive. Toward efficient adaptation from pre-trained LLMs to downstream tasks, several parameter-efficient fine-tuning methods have been proposed \cite{houlsby2019parameter,ben-zaken-etal-2022-bitfit,li2021prefix,hu2021lora,zhang2023adaptive,dettmers2024qlora}, which can achieve the performance close to or even surpass that of full fine-tuning with lower costs. 
\vspace{0.2cm}\\
Although these methods have demonstrated their effectiveness to some extent and have been widely applied, the underlying principles are still unclear and the selection of trainable modules are typically heuristically. \citet{li2018measuring} points out that over-parameterized networks have a lower intrinsic dimension, i.e. they can achieve 90\% or even 100\% of the full parameter performance by training only fewer parameters than the full parameter, which is the motivation of LoRA \cite{hu2021lora}. A natural question is:
\begin{center}
\vskip -0.1in
\textbf{\textit{What causes pre-trained language models to have the lower intrinsic dimension?}}
\end{center}
\vskip -0.1in
To answer this question, we first need to understand what changes occur from random initialization to pre-trained initialization. As indicated in \citet{devlin2018bert}, pre-trained models rapidly generalize to downstream data through fine-tuning. Intuitively, the generalization error of pre-trained models is controlled by a tighter bound compared to random initialized models. Therefore, we shed a spotlight on a classic PAC-Bayesian generalization error bound to analysis the difference between training from scratch and training after pre-training, which is first proposed by \citet{mcallester2003pac} and refined by \citet{maurer2004note,catoni2007pac,thiemann2017strongly}. In contrast to general PAC bounds that only consider the complexity of the hypothesis space, typically measured by VC-dimension \cite{vapnik1991principles} or Rademacher complexity \cite{bartlett2002rademacher}, PAC-Bayesian bounds jointly consider a less accurate prior probability distribution for parameters and a more precise data-based posterior distribution. We view pre-training as a shift of prior, where pre-training, by learning language features such as grammar and syntax on a large corpus, moves the model away from sub-manifolds of non-expressive parameters in the parameter space. Pre-training initialization is equivalent to assigning relatively low probabilities in the prior for non-expressive parameters. In contrast to random initialization, which assigns equal probabilities in the prior to all parameters, the KL divergence between the pre-trained prior and the data-based posterior is lower, as shown in Figure \ref{prior_posterior}. Therefore, pre-trained models have a tighter generalization error bound. We validate this prior shift through investigating changes in the loss landscape and gradient distribution. From random initialization to pre-trained initialization, the loss landscape transitions from flat oscillations to sharp oscillations, and the gradient exhibits a quasi-sparse property, where partial components dominate the majority of the gradient norm. Due to this difference, we conclude that \textbf{the intrinsic dimensions actually reflects the required dimensions of the search space}. For pre-trained models, the searching space dimensions needed to reach an optimal solution during the fine-tuning process has been compressed and the subspace spanned by the dimensions with extreme gradients largely contains the optima. As a result, the pre-trained models can be effectively adapted to downstream tasks by updating only a small portion of parameters.
\vspace{0.2cm}\\
Based on this, we propose a gradient-based sparse update scheme called \textbf{S}parse \textbf{I}ncrement \textbf{F}ine-\textbf{T}uning (SIFT). Unlike other existing PEFT methods, such as LoRA \cite{hu2021lora}, Adapter \cite{houlsby2019parameter}, Prefix-Tuning \cite{li2021prefix}, which require pluging in additional modules, SIFT does not alter the model's initial structure and is the first to achieve parameter efficiency in a component-sparse way. To achieve memory efficiency, it inserts hook functions into back propagation to acquire sparse gradients and perform in-place sparse updates on the parameters, as illustrated in Figure \ref{implementation}. We validate its effectiveness on a range of tasks including the GLUE Benchmark and Instruction-tuning and highlight its advantages in parameter efficiency in Section \ref{sparse_rate}. Overall, we demonstrates the feasibility of fine-tuning language models entirely in a component-sparse way.
\vspace{0.2cm}\\
In summary, the contributions of this paper are as follows:
\begin{itemize}
    \item To our best knowledge, this is the first time analyzing pre-trained models with PAC-Bayesian bounds. We validate the prior shift of pre-trained models from the perspectives of loss landscapes and gradient distributions and relate the low intrinsic dimensions to the compressed searching space.
    \item We propose a gradient-based sparse fine-tuning algorithm, an efficient and simple PEFT method, which is an entirely component-sparse method that needs no additional plug-in modules. It demonstrates competitive performances and better parameter utilization efficiency.
    \item We introduce a memory-efficient implementation for SIFT by utilizing hooks during backward propagation, which could potentially be implemented at the lower level of deep learning training frameworks like PyTorch \cite{paszke2017automatic}, and integrated into the general training process. This paves the way for more streamlined and efficient fine-tuning.
\end{itemize}

\section{Related Work}

\textbf{Parameter-efficient Fine-tuning:}
To apply pre-trained models more efficiently to downstream tasks, \citet{houlsby2019parameter} proposes parameter-efficient transfer learning by inserting a bottleneck structure between different layers of the pre-trained model. During fine-tuning, the pre-trained parameters remain unchanged, and only the bottleneck layers are updated. \citet{ben-zaken-etal-2022-bitfit} shows that training only bias terms and task-specific modules during fine-tuning can also adapt pre-trained models effectively to downstream tasks. Prefix-tuning \cite{li2021prefix} adapts pre-trained models to downstream tasks by prepending trainable continuous tokens (Prefixes) to the input and hidden states of each Transformer layer. LoRA \cite{hu2021lora} freezes the weights of pre-trained models and injects trainable rank decomposition matrices into each layer of the Transformer architecture. Following the proposal of LoRA, several LoRA-based improvements have been introduced, including AdaLoRA \cite{zhang2023adaptive}, which adaptively allocates parameter budgets based on the importance of weight matrices, and QLoRA \cite{dettmers2024qlora}, which further reduces the resource overhead through quantization techniques. The above PEFT methods, except for Bias-only which is a module sparse approach, are all adapted to the downstream tasks by introducing additional modules.
\vspace{0.2cm}\\
\textbf{PAC-Bayesian Generalization Error Bounds:} introduced by \citet{mcallester1999pac,mcallester2003pac}, combines the advantage of Bayesian methods and PAC Learning, which allows for considering domain knowledge as a Bayesian prior without assuming the truth of the prior \cite{mcallester2003pac}. Early generalization error bounds only consider the complexity of the hypothesis space, measured by VC-dimension \cite{vapnik1991principles} and Rademacher complexity \cite{bartlett2002rademacher}, which is nearly vacuous and trivial in the real neural network setting \cite{zhang2021understanding}. In contrast, PAC-Bayes methods reward models whose prior assigns higher probability to those hypotheses that are in line with train data, with a tighter control over generalization error.

\section{Understanding Pre-training-fine-tuning from a Distribution-shift Perspective  }
The pre-training-fine-tuning paradigm has shown great success in numerous NLP tasks \cite{devlin2018bert}. In this section, we adopt the PAC-Bayesian generalization error bounds and, from the perspective of the prior distribution shift, analyze the pre-training-fine-tuning paradigm in both theoretical and empirical aspects. We conclude that the low intrinsic dimensions of pre-trained models originate from the compressed searching space and the subspace spanned by the dimensions with extreme gradients largely contains an optima.

\subsection{PAC-Bayesian Generalization Error Bounds}
For a fixed architecture of the deep learning model, models with different parameter values (i.e., hypotheses) collectively form the hypothesis space. Any hypothesis $h$ in the hypothesis space $\mathcal{H}$ can be represented as $h=\sum_{i=1}^{N}h_i e_i$, where $N$ is the total number of parameters, $e_i$ is akin to the definition of the standard basis (with the $i^{th}$ component being 1 and the rest being 0), and $h_i$ represents the value of the model $h$ on the $i^{th}$ component. Hence, a straightforward conclusion is that $|\mathcal{H}|\leq N$, and subsequent discussions are based on the premise that the hypothesis space is finite. 
\vspace{0.2cm}\\
Let the empirical risk of $h$ be $\hat{\mathcal{R}}(h)=\frac{1}{n}\sum_{i=1}^{n}l(h,s_i)$, and the population risk be $\mathcal{R}(h)=\underset{s\sim\mathcal{S}}{\mathbb{E}}[l(h,s)]$, where $s_i$ is independently sampled from the identical data distribution $\mathcal{S}$, $n$ represents the number of samples, $l$ is a bounded loss function, and for simplicity, $l(h,s)\in[0,1]$. The generalization error is defined as $\Delta(h)=\hat{\mathcal{R}}(h)-\mathcal{R}(h)$.
\vspace{0.2cm}\\
In this setting, with a probability at least $1-\delta$, the population risk $\hat{\mathcal{R}}(h)$ satisfies:
\begin{equation}\label{eq:ieq1}
    \forall h \in \mathcal{H},\hat{\mathcal{R}}(h)\leq\mathcal{R}(h)+\sqrt{\frac{log(|\mathcal{H}|)+log(1/\delta)}{2n}}
\end{equation}
If we consider a probability distribution $\mathcal{P}$ regarding hypotheses, referred to as a prior \cite{mcallester1998some}, then with a probability of at least $1-\delta$,
\begin{equation}\label{eq:ieq2}
\forall h \in \mathcal{H}, \hat{\mathcal{R}}(h) \leq \mathcal{R}(h)+\sqrt{\frac{\log(1/\mathcal{P}(h)) + \log(1/\delta)}{2n}}
\end{equation}
A most trivial prior is to assume that all hypotheses are equally likely, i.e., $\forall h \in \mathcal H, \mathcal{P}(h) = 1/\log(|\mathcal{H}|)$. In this case, the inequality \ref{eq:ieq2} degenerates to \ref{eq:ieq1}. But the parameters are certainly not uniformly distributed. For natural language processing tasks, without considering specific data, parameters that can accurately represent fundamental language features such as grammar and semantics should be assigned a higher probability. We refer to this as a weakly informative prior, which conveys partial information.
\vspace{0.2cm}\\
If we consider a more precise distribution $\mathcal{Q}$, which characterizes the probability distribution of the parameters based on specific data, referred to as a posterior, then we have the following bound on the population risk, first introduced by \citet{mcallester2003pac}, with a probability at least $1-\delta$:
\begin{equation}\label{eq:ieq3}
\underset{h \sim \mathcal{Q}}{\mathbb{E}}[R(h)] \leq \underset{h \sim \mathcal{Q}}{\mathbb{E}}[\hat{R}(h)] + \sqrt{\frac{\mathbb{KL}(Q||P)+\log({n/\delta})+2}{2n-1}}
\end{equation}

\begin{figure}[t]
\begin{center}
\includegraphics[width=0.4\textwidth]{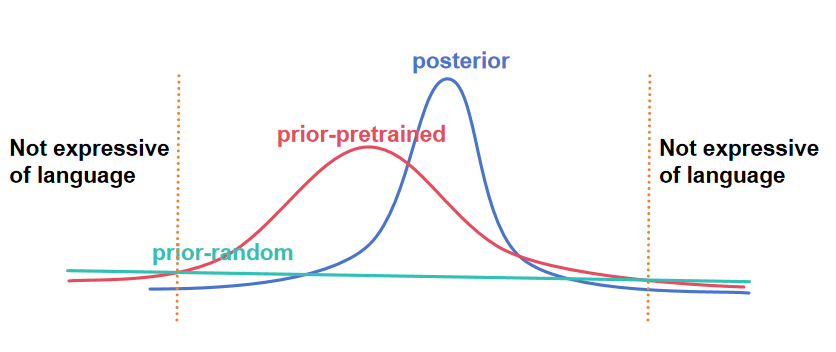}
\caption{An intuitive explanation about prior distribution and posterior distribution. Random initialization is equivalent to assigning equal probabilities to hypotheses in the hypothesis space (the horizontal line referred as prior-random). Pretraining learns language features from extensive corpora, moving away from hypotheses that are not accurately expressive of language, equivalent to assigning them lower prior probabilities. Data-based, more precise posterior also have lower probabilities for hypotheses that fail to represent and understand language correctly. Therefore, compared to prior-random, the KL divergence between prior-pretrained and posterior is smaller.}
\label{prior_posterior}
\end{center}
\vskip -0.2in
\end{figure}

The formula \ref{eq:ieq3} indicates that the smaller KL divergence between the prior and the posterior, the tighter control there is on the generalization error. A more detailed introduction of the above theorems can be found in Appendix \ref{appendix-pac-bayes}.
\vspace{0.2cm}\\
Pre-training language models aims to learn grammar, semantics, and other basic language features (referred to as a prior) from a large corpus through masking or auto-regressive methods, steering the model away from parameters that cannot correctly understanding and convey language. Therefore, pre-trained initialization of the model is equivalent to assigning lower probabilities to such parameters in the prior. Those parameters are naturally incapable of handling various tasks related to language correctly, so they should also have lower probabilities in the data-based posterior. As shown intuitively in Figure \ref{prior_posterior}, pre-trained initialization considering basic language features is closer to the distribution of the data-based on  posterior compared to random initialization. This closeness results in lower KL divergence and, consequently, a tighter upper bound on generalization error, allowing the population risk to consistently decrease along with the reduction of empirical risk during training.
\vspace{0.2cm}\\
The difference in the distribution of parameters can significantly affect the variation of losses, i.e., the properties of loss landscapes. In Section \ref{sec-loss-landscape}, we validate this distribution difference between random and pre-trained initialization by visualizing the loss landscape. By the first-order approximation of the multivariate Taylor expansion, this difference in loss landscapes stems from varied gradient distributions, which is also confirmed by the empirical analysis in Section \ref{sec-grad-dist}.

\begin{figure*}[t]
\begin{center}
\includegraphics[width=0.20\textwidth]{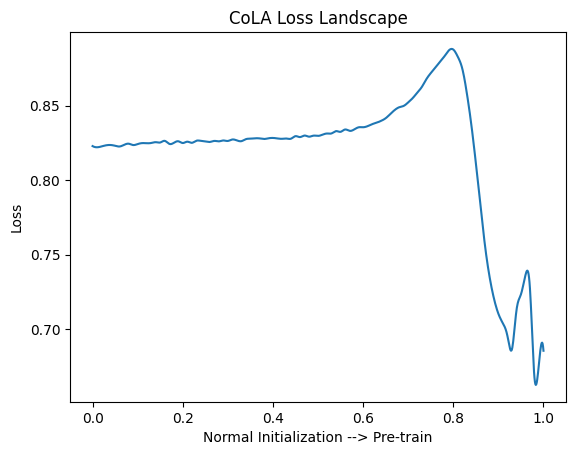}
\includegraphics[width=0.20\textwidth]{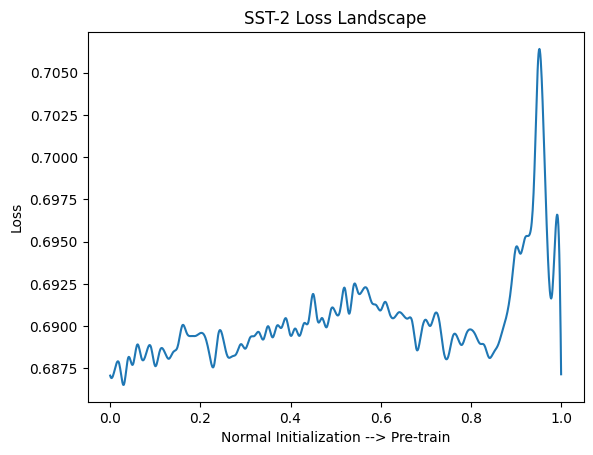}
\includegraphics[width=0.20\textwidth]{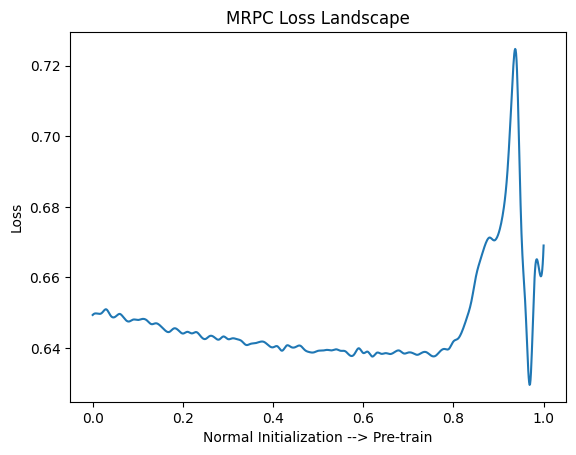}
\includegraphics[width=0.20\textwidth]{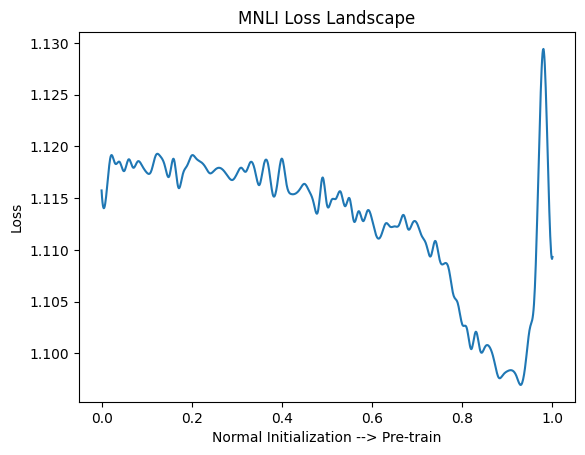}
\includegraphics[width=0.20\textwidth]{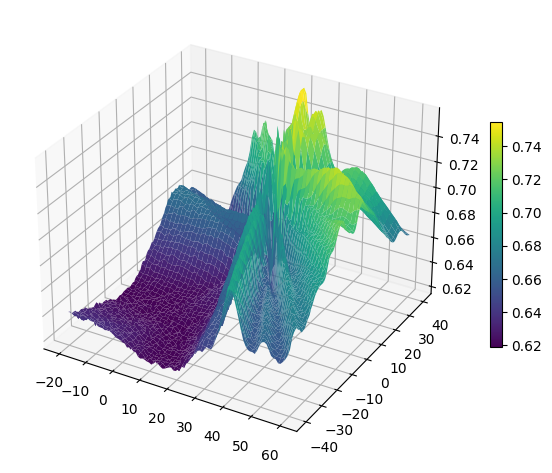}
\includegraphics[width=0.20\textwidth]{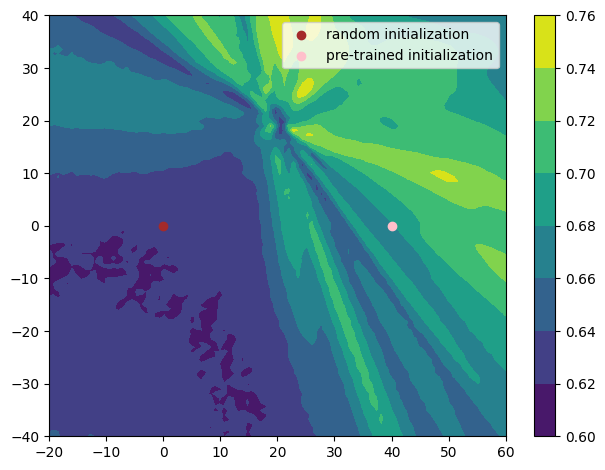}
\includegraphics[width=0.20\textwidth]{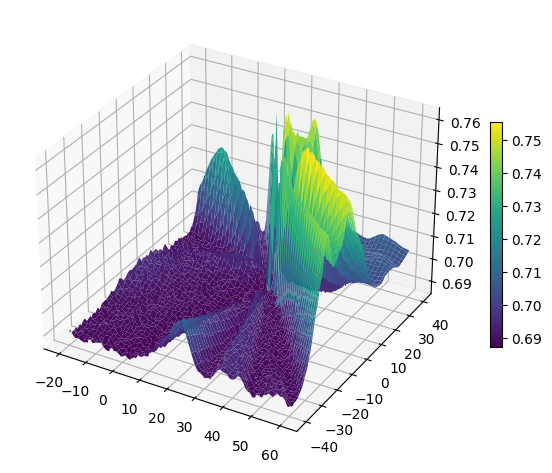}
\includegraphics[width=0.20\textwidth]{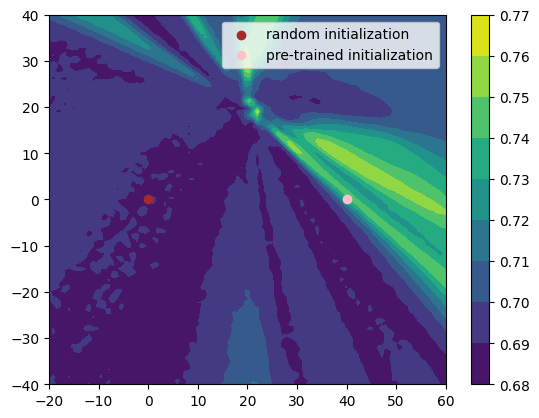}
\caption{1-D (up) and 2-D (down) visualization of the shift of the loss landscape from random initialization to pre-trained initialization. The figures show the loss landscape transitions from low amplitude oscillations to high amplitude oscillations. }
\label{losslandscape}
\end{center}
\vskip -0.2in
\end{figure*}

\subsection{Visualization of Loss Landscape}\label{sec-loss-landscape}
Loss landscape is a high-dimensional geometry which is completely determined as soon as a dateset and the network structure are specified \cite{li2018measuring}, reflecting the relationship between the loss and parameters. For high-dimensional geometries such as loss landscapes, it is not possible to imagine and understand their geometric properties as intuitively as in low-dimensional shapes. Despite the difficulty in understanding such high-dimensional geometries, researchers have proposed methods to project high-dimensional loss surfaces into low-dimensional space, including one-dimensional visualization \cite{goodfellow2014qualitatively,Smith2017ExploringLF} and two-dimensional visualization \cite{li2018visualizing}, in order to easily understand and study some of the properties of loss landscapes. 
\vspace{0.2cm}\\
In exploring the shift of the loss landscape from random initialization to pre-trained initialization, we note $\theta_0$ as the randomly initialized parameters and $\theta_1$ as the pre-trained parameters. \citet{goodfellow2014qualitatively} plots the loss landscape curve from $\theta_0$ to $\theta_1$ by linearly sampling and interpolating the network parameters between $\theta_0$ and $\theta_1$, the loss curve $f$ is defined as:
$$\begin{aligned}
    f(\alpha) 
    &=\mathcal{L}(\theta_0 + \alpha \delta)
\end{aligned}$$
where $\alpha \in [0, 1]$ and $\delta$ is $\theta_1 - \theta_0$ that represents a direction.
The two-dimensional loss surface $f$ is plotted similarly to the one-dimensional one, determined by a starting point and two directions:$$f(\alpha, \beta)=\mathcal{L}(\theta_0 + \alpha \delta_1 + \beta \delta_2). $$In our setting, $\delta_1=\theta_1 - \theta_0, \alpha \in [-0.5, 1.5], \beta \in [-1, 1]$, when $\alpha=0, \beta=0$, the model is in the random initialization state, and when $\alpha=1, \beta=0$, the model is in the pre-trained state. We use a method similar to \citet{li2018visualizing} to construct the direction $\delta_2$ to avoid the scaling effect between different components, multiplying $\delta_1$ by a Gaussian random number points by points, i.e. $\delta_{2_{i}} = d_i * \delta_{1_i},$ where $d_i$ follows a normal distribution. We investigate the shift of the loss landscape using RoBERTa-base \cite{liu2019roberta} on a range of tasks at GLUE, which are shown in Figure \labelcref{losslandscape}.
\vspace{0.2cm}\\
A prior idea would be that pre-trained models are more efficient at adapting to downstream tasks than training from scratch because the knowledge learned during pre-training reduces the initial loss. However, a striking fact is that pre-training does not make all tasks have lower losses than random initialization at the beginning of fine-tuning, e.g. MNLI, MRPC, etc. Instead, for most task, there are extremely higher losses than random initialization around the pre-trained initialization. A common feature of the loss landscape is that the oscillations of the loss around random initialization are generally small in magnitude, whereas the loss oscillations around pre-training are sharp, i.e. after a valid pre-training process, the loss landscape shifts from low amplitude oscillations to high amplitude oscillations.
\vspace{0.2cm}\\
Since the 1-D and 2-D loss landscapes are projections of the original loss landscape, the special states exhibited in the low-dimensional projections reflect the general property of the high-dimensional space. For simplicity, let's consider the one-dimensional loss curve, $L(\theta_0 + \alpha \delta)$, where $\delta$ can be represented as a linear combination of natural bases, i.e. $\delta=\sum_{i=1}^{N} d_i e_i$. By the first-order approximation of the multivariate Taylor expansion, we can estimate 
\begin{align*}
    L(\theta_0 + \alpha \delta)&=L(\theta_0 + \alpha \sum d_i e_i)\\ 
                                &\approx L(\theta_0) + \alpha \sum d_i \nabla L_{e_i}(\theta_0)
\end{align*}

\begin{figure*}[t]
\begin{center}
\includegraphics[width=0.33\textwidth]{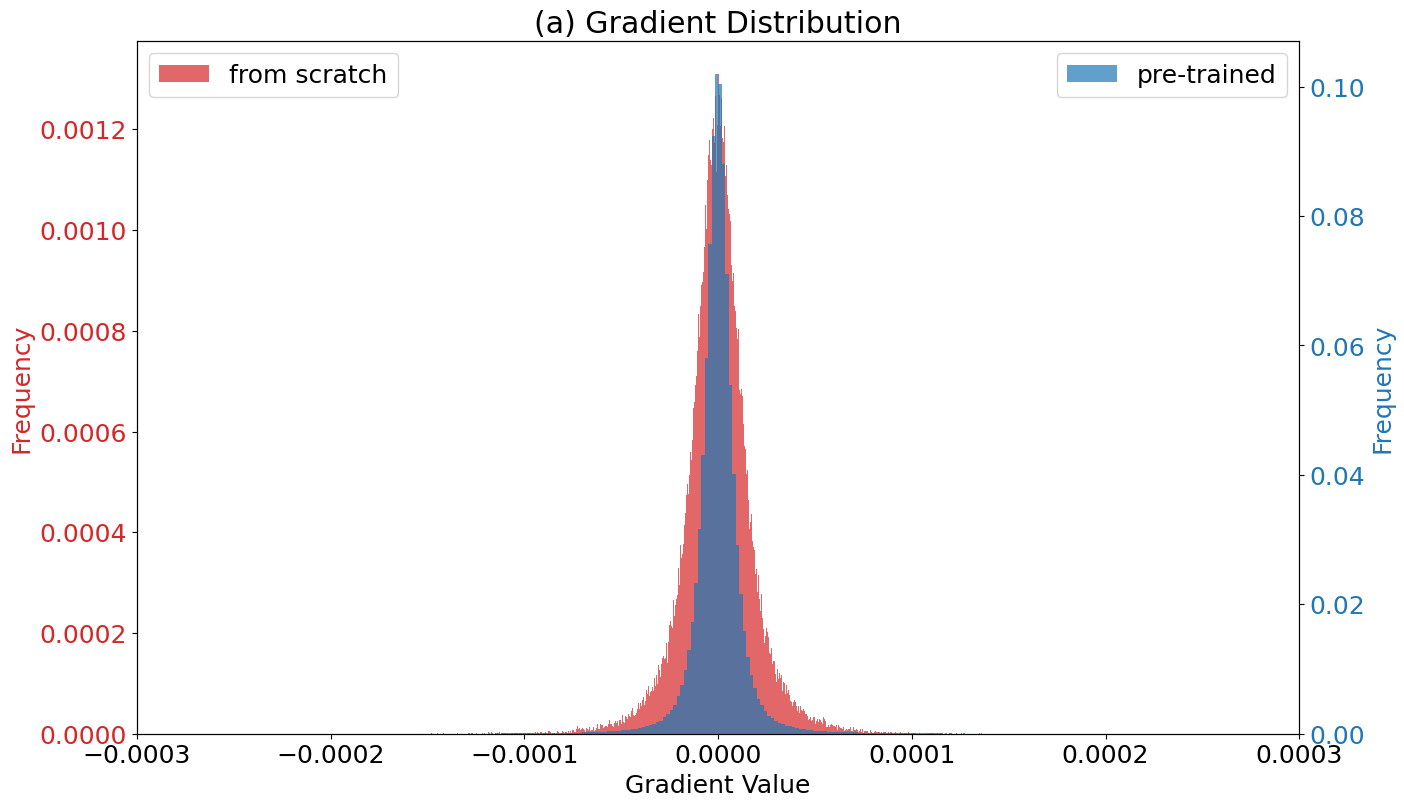}
\includegraphics[width=0.33\textwidth]{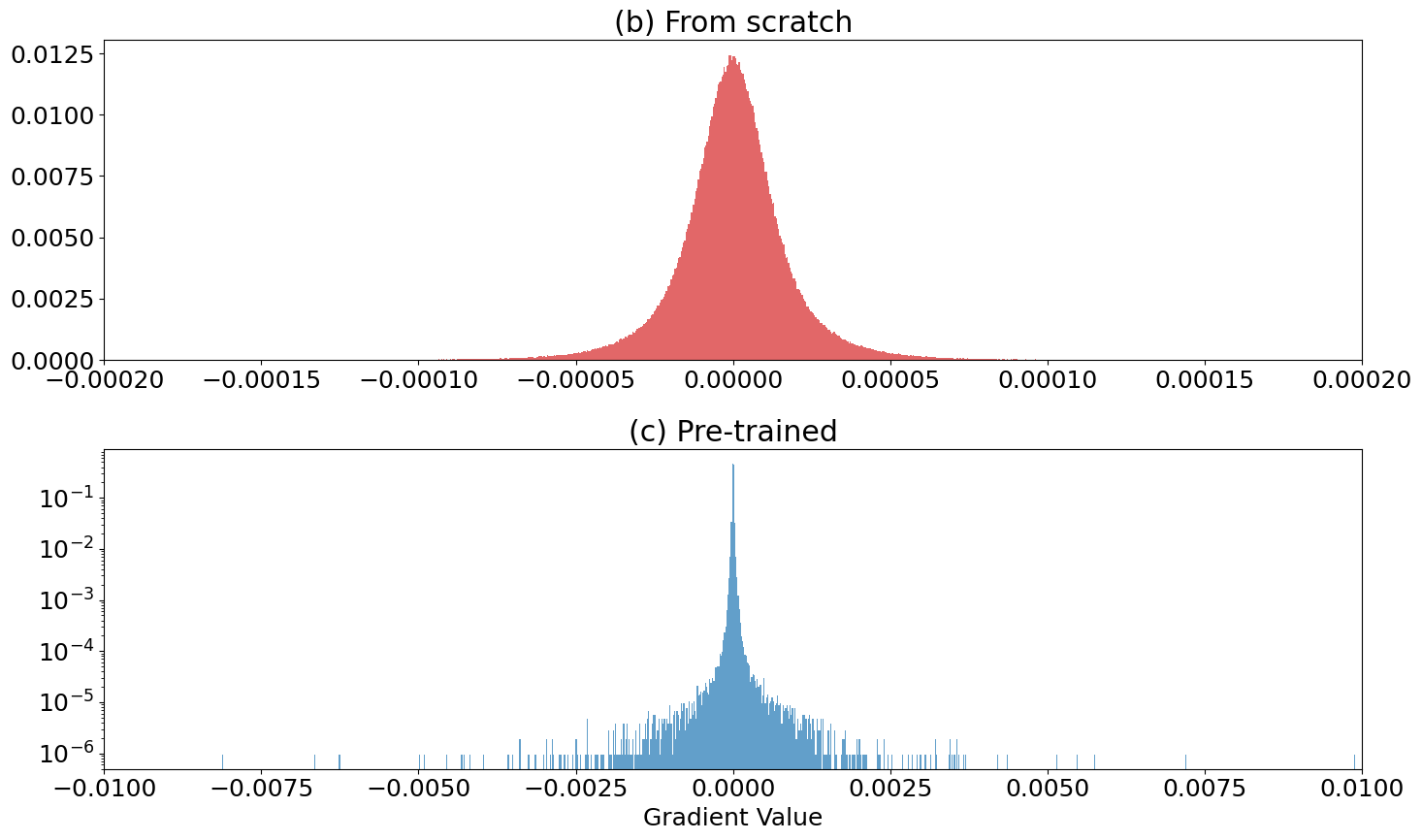}
\includegraphics[width=0.33\textwidth]{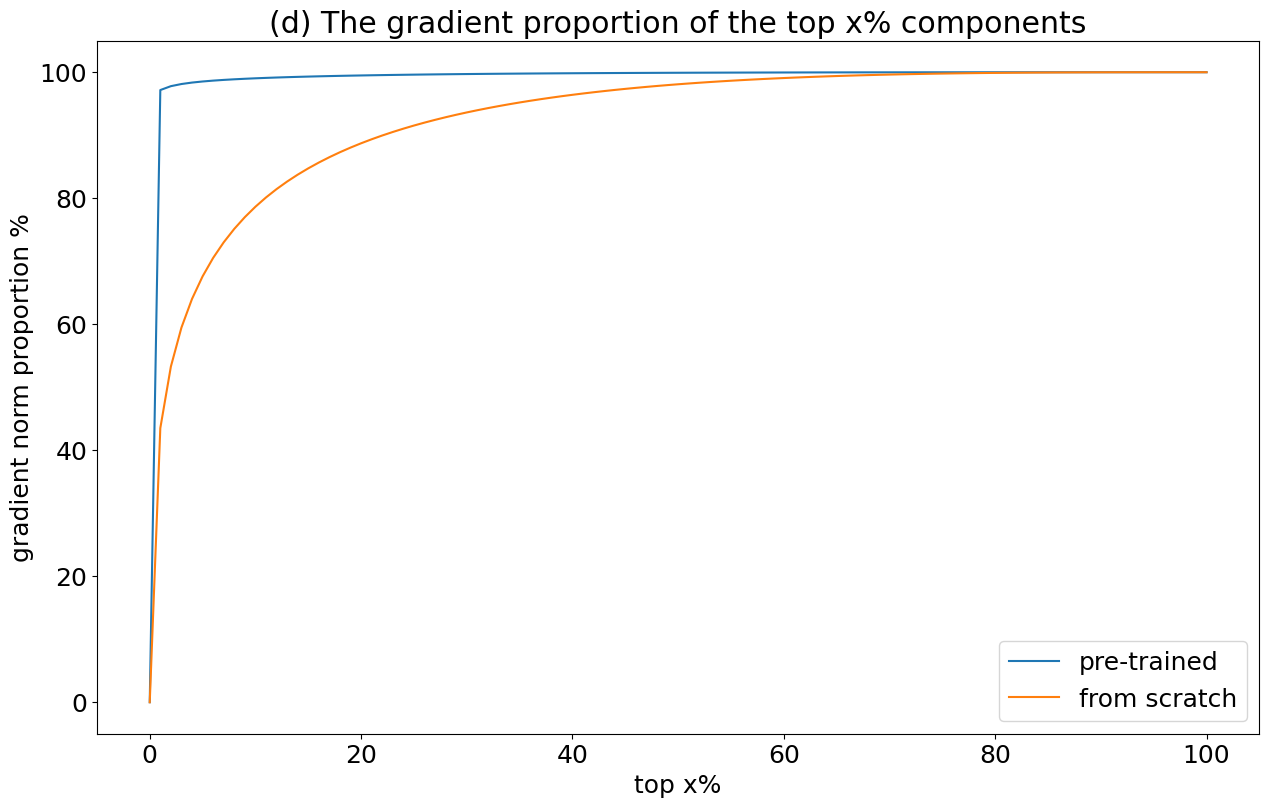}
\caption{(a) depicts the gradient distribution of parameters in one layer of RoBERTa-Large on the MNLI dataset, showing different gradient distributions of the models trained from scratch and from pre-trained. Both distribution are approximately presenting a bell-shaped distribution similar to a normal distribution with a mean of zero. (b) and (c) are drawn on their respective scales, showing that compared to the model trained from scratch, the gradients of the pre-trained model are more concentrated around zero while they exists much larger gradient values. (d) illustrates the gradient proportion of the top x\% components, indicating that the gradients of the pre-trained model hold a more extreme property, dominating 99\% of the complete gradient norm with only 1\% of the parameters.}
\label{grad_dist}
\end{center}
\vskip -0.2in
\end{figure*}
If the probability distributions of gradients on each dimension $e_i$ are relatively similar, meaning there is an expectation $E(\nabla L_{e_i}(\theta_0))\approx E_0$, where $E_0$ is a constant, and since $N$ is usually quite large, the loss can be further approximated as $L(\theta_0) + \alpha \sum d_i E_0$. This indicates that the loss will not exhibit sharp oscillations. However, if extreme distributions exist in certain dimensions, for example, if the $N^{th}$ dimension has a different distribution, then the loss can be approximated as $L(\theta_0) + \alpha \sum_{i=1}^{N-1} d_i E_0+ \alpha d_N \nabla L_{e_N}(\theta_0)$. In this case, the loss will be significantly influenced by the gradient of the $N^{th}$ dimension. Therefore, a flat loss landscape implies similar distributions among dimensions and the intense oscillations in loss landscape indicate that certain dimensions dominate the changes of the loss, suggesting an inconsistency in the distributions among dimensions. We verify this inconsistency in gradient distributions in the following Section \ref{sec-grad-dist} and highlight the quasi-sparsity in gradients of pre-trained models.

\subsection{Quasi-Sparse Gradient Distribution}\label{sec-grad-dist}

\citet{ye2020accelerating,wiedemann2020dithered} take the assumption that the distribution of the parameter gradients approximates a normal distribution. In Figure \labelcref{grad_dist}, we plot the gradient distribution of one parameter in RoBERTa-Large on the MNLI dataset and other parameters also have the similar property shown in the following part. Both the gradients of the model trained from scratch and from pre-trained show a bell-shaped distribution with a mean of near zero. Under the same histogram settings, the gradient distributions of from scratch and pre-trained models show a huge difference in frequency and value range. Figure \ref{grad_dist}(c) displays the pre-trained gradient distribution on a log-scale, where the majority of gradients are concentrated around zero, with only a very small portion having larger gradient values, and the maximum value of pre-trained is about two orders of magnitude higher than that from scratch. Figure \ref{grad_dist}(d) shows the gradient proportion of the components with the top x\% absolute gradient value. The pre-trained gradient shows a more extreme property, where 1\% of the components account for 99\% of the total gradient norm. This severe imbalance in distribution exhibits a property similar to sparsity, which we refer to as Quasi-Sparse. The majority of information in the gradient space can be represented by a very small number of components.
\vspace{0.2cm}\\
Through pre-training, the model reaches a sub-manifold in the parameter space where the parameters are with higher distribution probabilities and can be expressive of language (corresponding to the bell-shaped distribution in Figure \ref{prior_posterior}). Therefore, during the fine-tuning process on downstream tasks, it is only necessary to search within this manifold to find a good solution. This results in redundancy on most dimensions, manifested by smaller gradient values. Therefore, the compressed searching space is the key to the low intrinsic dimension of the objective loss landscape \cite{li2018measuring}. A natural question is whether updating only in these dimensions can effectively adapt the pre-trained model to downstream domains. A prior idea is that, due to the quasi-sparsity of the gradient, updating only partial components with larger gradients can also effectively reduce the empirical loss, while the generalization loss can also be consistently reduced due to tighter upper bounds on the generalization error of pre-trained models.
\vspace{0.2cm}\\
In the following Section \ref{sec-method}, we propose a memory-efficient implementation of the sparse fine-tuning scheme, which validates the feasibility of fine-tuning language models entirely in a component-sparse way.


\section{Methodology}\label{sec-method}
We describe the design of Sparse Increment Fine-Tuning (SIFT), whose principle is to update only a part of the components of the parameters. Based on the similar subspace method in nonlinear optimization\cite{CSIAM-AM-2-585}, we provide a mathematical definition related to SIFT. Finally, we propose a memory-efficient implementation for SIFT.

\subsection{SIFT: Sparse Increment Fine-Tuning}

We define the fine-tuned parameters as the sum of pre-trained parameters and an increment, 
$x_{ft}=x_{pt}+\Delta x.$
Updating only a portion of the components means that $\Delta x$ is a sparse matrix, hence we name this update method Sparse Increment Fine-Tuning (SIFT). Below, we will explain that due to the quasi-sparse gradient property of the pre-trained model, $\Delta x$ can ensure sufficient descent of the loss.
\vspace{0.2cm}\\
The idea of only updating partial components of the parameters is similar to the subspace method for non-linear optimization \cite{CSIAM-AM-2-585}. In SIFT, we aim to update only components with the top x\% absolute values of the gradient, so the update direction will be in a subspace spanned by these components’ dimensions. We adopt the definition in \cite{CSIAM-AM-2-585}, where $g_k^i$ is denoted as the $i^{th}$ component of gradient $g_k$. Sorted in descending order of absolute value, the top x\% components are selected:
$$\begin{aligned}
    &|{g_k}^{i_1}| \geq |{g_k}^{i_2}| \geq |{g_k}^{i_3}| \geq \cdots \geq |{g_k}^{i_m}|,\\ 
    &m= [x\% * n],\\
    &n\ is\ the\ number\ of\ g_k \ components 
\end{aligned}$$
Therefore, the update direction $d$ are in the subspace $\mathcal{S}= \text{span} \{e^{i_1}, e^{i_2}, \ldots, e^{i_n}\}$, called the $\tau$-steepest coordinates subspace, where $e_i$ refers to the $i^{th}$ standard basis. Based on this, we define the steepest descent direction as sufficiently descending, when,
$$\min_{d \in \mathcal{S}} \frac{d^T g_k}{{\|d\|}_2{\|g_k\|}_2} \leq -\frac{\tau}{n}$$
That is, the descent direction d is sufficiently close to the negative gradient direction. If ${(g_k^{\tau+1})}^2 \leq \epsilon \sum_{j=1}^{\tau} {(g_k^j)}^2$, we can further obtain the following estimate,
$$\min_{d \in \mathcal{S}} \frac{d^T g_k}{{\|d\|}_2{\|g_k\|}_2} \leq -\frac{1}{\sqrt{1+\epsilon(n-\tau)}}$$
As discussed in section 3.2, for the gradients of pre-trained models, a few components dominate most of the gradient norm, allowing us to obtain a sufficiently small $\epsilon$ with a relatively small $\tau$, ensuring that the direction d chosen from the subspace $\mathcal{S}$ is sufficiently descending.

\begin{figure}[t]
\begin{center}
\includegraphics[width=0.23\textwidth]{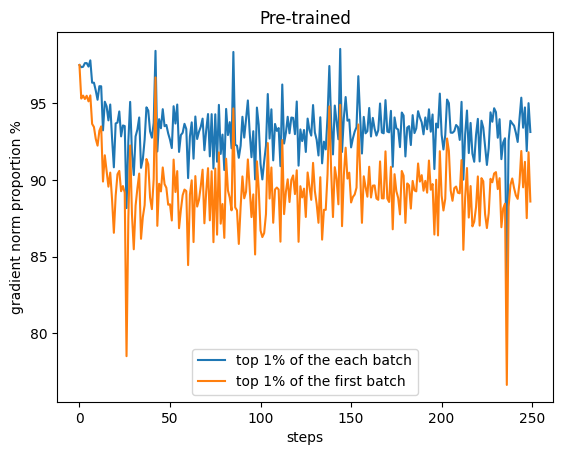}
\includegraphics[width=0.23\textwidth]{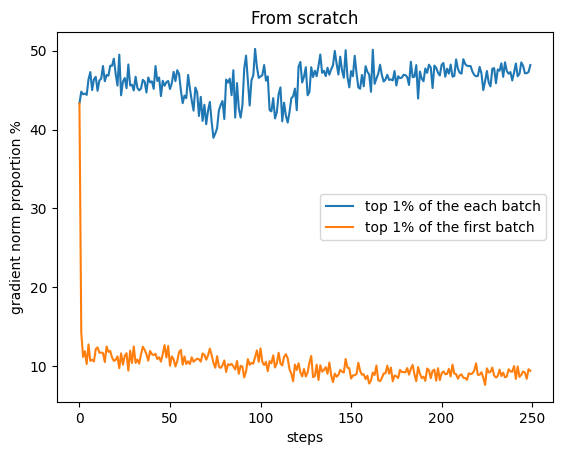}
\caption{(Left) compares the difference in the proportion of gradients accounted for by two different top 1\% component selection methods: the top 1\% of the current batch and the top 1\% of the first batch. Although sticking with the first batch's top 1\% as opposed to selecting the top 1\% of the current batch results in some reductions of gradient information, the difference remains within an acceptable range. Avoiding frequent changes of components can improve training efficiency and preserve the historical information of Adam-like optimizers. In contrast, (Right) shows the gradient proportion variation in the model trained from scratch, where fixing the top 1\% of the first batch leads to a greater reduction of the gradient information.}
\label{grad_switch}
\end{center}
\vskip -0.3in
\end{figure}

Due to the limitation of computational cost, we usually cannot obtain the gradients over the complete data. Instead, stochastic gradient type methods typically take a mini-batch sampled from the full data. Without the complete gradient information, SIFT uses the gradient from the first batch(or first several batches by using gradient accumulation) as an estimate of the complete gradient and selects the components to be updated based on this. A natural question arises: do the components selected based on the gradient of the first batch still effectively represent the full gradient information in subsequent batches? That is, whether there will be significant differences between the gradients of different batches. As shown in Figure \labelcref{grad_switch}, we compare the proportion of the top 1\% components of each batch's gradients with the proportion of the top 1\% components of the first batch's gradients. For pre-trained models, using the top x\% components of the first batch does indeed result in some reduction of gradient information, but except for some outlier samples, the difference always remains within an acceptable level compared to the top 1\% of the current batch. In contrast, the top 1\% components of the first batch in models trained from scratch tend to lose more gradient information. Therefore, for efficiency, SIFT always updates the top x\% components of the first batch in the whole training process. This approach enables the use of Adam-like optimizers that require historical state information, as frequent changes of components would result in the loss of their historical information. Appropriate periodic changes of the components that need to be updated is also an acceptable choice, and this could be one of the focus for future work.

\subsection{A Memory-efficient Implementation of SIFT}

The key to SIFT lies in obtaining gradients of partial components. Under existing deep learning frameworks (such as PyTorch \cite{paszke2017automatic}), gradients of all parameters are able to be accessed only after backward propagation. A simple implementation of SIFT is to retrieve the gradients of partial components through indexing after all gradients have been calculated. However, as all gradients have already been stored, storing the gradients of partial components at this point does not reduce the memory overhead of gradient storage. The memory overhead during training mainly consists of four parts: parameters, gradients, optimizer states, and activation. Although this implementation does not reduce storage overhead of gradients, it can significantly reduce that of the optimizer states.
\begin{figure}[ht]
\begin{center}
\includegraphics[width=0.4\textwidth]{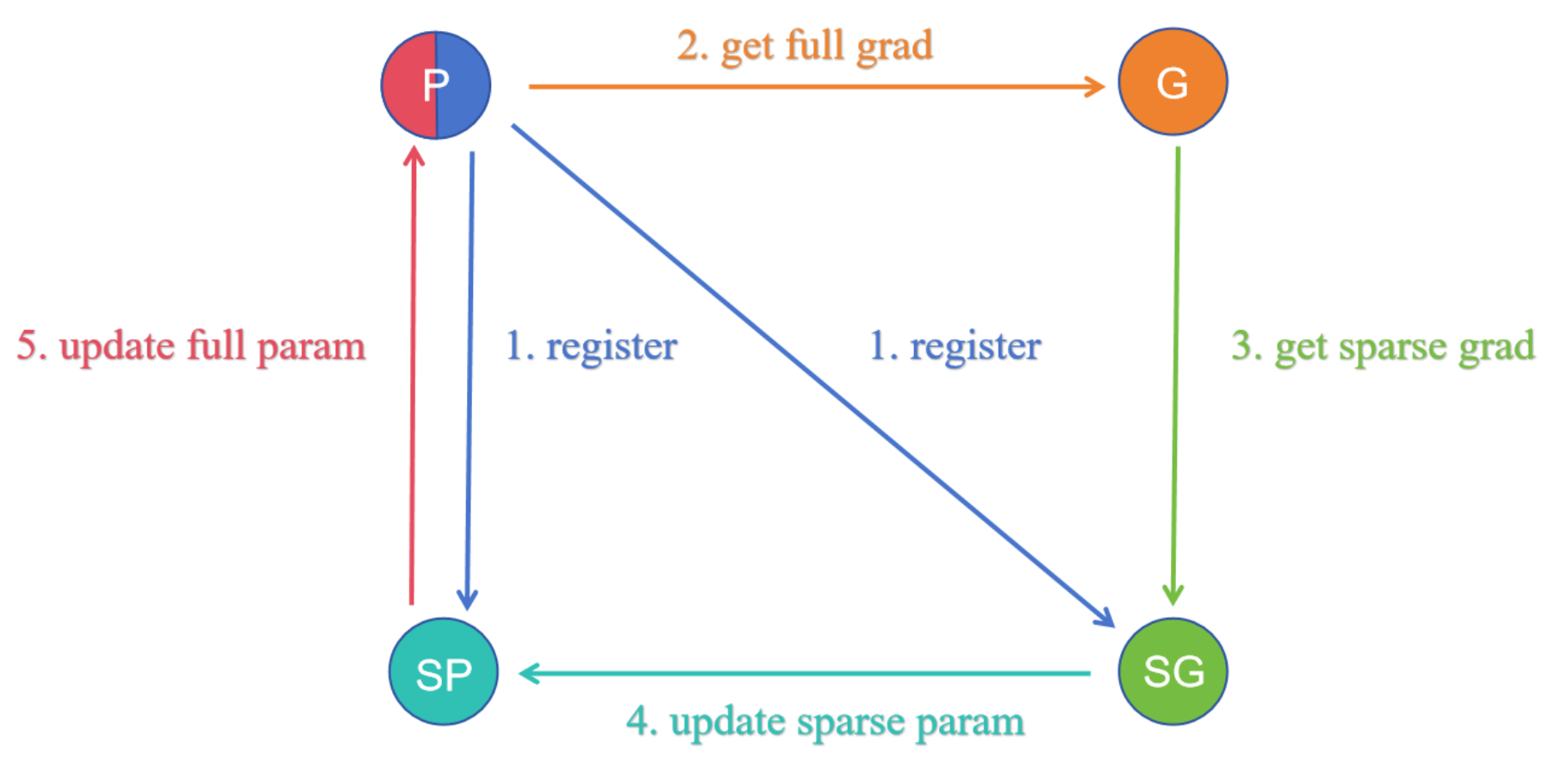}
\caption{(1) For each parameter (P) that requires updating sparsely, we register a Sparse Parameter (SP) with the Sparse Gradient (SG), storing them in the form of values and indexes; (2) Acquire the  computed gradient (G) by inserting a hook function; (3) Obtain partial components of the gradient through indexing to serve as the Sparse Gradient for the Sparse Parameter; (4) Use the Sparse Gradient to update the Sparse Parameter in the optimizer; (5) Use the Sparse Parameter to update the initial parameter.}
\label{implementation}
\end{center}
\vskip -0.2in
\end{figure}

\begin{table*}[t]
\caption{Performances of fine-tuned pre-trained RoBERTa-large with different methods on the GLUE benchmark. Num denotes the number of trainable parameters in the fine-tuning process, excluding the classifier. Similar to \cite{hu2021lora}, we report the matched and mismatched accuracy for MNLI, Matthew’s correlation for CoLA, Pearson correlation for STS-B, and accuracy for other tasks. * indicates results reported in \cite{hu2021lora}. The best results are highlighted in \textbf{bold} and the second best results are \underline{underlined}.}
\label{glue}
\vskip 0.15in
\begin{center}
\begin{small}
\begin{sc}
\renewcommand{\tabcolsep}{3pt}
\begin{tabular}{ccrccccccccc}
\toprule
Model & Method & Num & MNLI & SST-2 & MRPC & CoLA & QNLI & QQP & RTE & STS-B & Avg. \\
\midrule

\multirow{7}{*}{RoBERTa-Large} & Full\textsuperscript{*} & 355.0M & 90.2 & \underline{96.4} & \textbf{90.9} & 68.0 & \underline{94.7} & \textbf{92.2} & \textbf{86.6} & \textbf{92.4} & \textbf{88.9} \\

& Adpt\textsuperscript{P}\textsuperscript{*} & 3.0M & 90.2 & 96.1 & 90.2 & \underline{68.3} & \textbf{94.8} & 91.9 & 83.8 & 92.1 & 88.4 \\

& Adpt\textsuperscript{P}\textsuperscript{*} & 0.8M & \underline{90.5} & \textbf{96.6} & 89.7 & 67.8 & \textbf{94.8} & 91.7 & 80.1 & 91.9 & 87.9 \\

& Adpt\textsuperscript{H}\textsuperscript{*} & 6.0M & 89.9 & 96.2 & 88.7 & 66.5 & \underline{94.7} & \underline{92.1} & 83.4 & 91.0 & 87.8 \\

& Adpt\textsuperscript{H}\textsuperscript{*} & 0.8M & 90.3 & 96.3 & 87.7 & 66.3 & \underline{94.7} & 91.5 & 72.9 & 91.5 & 86.4 \\

& LoRA\textsuperscript{*} & 0.8M & \textbf{90.6} & 96.2 & 90.2 & 68.2 & \textbf{94.8} & 91.6 & 85.2 & \underline{92.3} & \underline{88.6} \\

& SIFT & 0.8M & \textbf{90.6} & 96.2 & \underline{90.4} & \textbf{68.5} & 94.1 & 91.0 & \underline{85.9} & \underline{92.3} & \underline{88.6} \\

\bottomrule
\end{tabular}
\end{sc}
\end{small}
\end{center}
\vskip -0.1in
\end{table*}
\citet{lv2023parameter} proposes an alternative SGD method that reduces gradient storage by inserting a hook function, thereby fusing the gradient into the parameters as soon as the gradient is computed. Inspired by \citet{lv2023parameter}, we propose a memory-efficient implementation of SIFT. As shown in Figure \labelcref{implementation}, we register a Sparse Parameter (SP) and corresponding Sparse Gradient (SG) for the parameters to be updated, only storing values and indexes for each. SP does not obtain SG through normal backward propagation. Instead, we inject a hook function into the parameters to capture the computed gradients and assign them to SG via indexing. SP with SG is updated through the normal optimizer, and finally, SP is integrated into the initial parameter P.
\begin{table}[h]
\caption{Memory consumption of fine-tuning Llama 7B with different methods including full fine-tuning and SIFT in 5\% sparsity rate. The data type of the model is bf16, the optimizers all use AdamW and are in fp32, and the sequence length and batch size are 2048 and 1. \textbf{GC} refers to whether gradient checkpointing is enabled or not.}
\label{Mem}
\begin{center}
\begin{tiny}
\begin{sc}
\begin{tabular}{ccccccc}
\toprule
& GC & Params & Grad & Optim States & Activation & Total \\
\midrule
\multirow{2}{*}{Full} &\XSolidBrush & \multirow{2}{*}{12.55} & \multirow{2}{*}{12.55} & \multirow{2}{*}{50.21} & 35.72 & 111.03\\

  & \CheckmarkBold & & & & 0.81 & 76.12\\
\midrule
\multirow{2}{*}{SIFT} &\XSolidBrush & \multirow{2}{*}{12.55} & \multirow{2}{*}{\textbf{0.626}} & \multirow{2}{*}{\textbf{2.51}} & 35.72 & \textbf{51.41}\\

  &\CheckmarkBold & & & & 0.81 & \textbf{16.50}\\
\bottomrule
\end{tabular}
\end{sc}
\end{tiny}
\end{center}
\vskip -0.1in
\end{table}

For x\% sparse updates, we can simultaneously reduce the gradients and optimizer states to the original x\%. Combined with techniques such as mixed-precision training and gradient checkpointing, it is possible to fine-tune a 7B model on a single RTX 3090 24GB, as shown in Table \ref{Mem}. 


\section{Experiments}
Our evaluation mainly consists of two parts, targeting two cases: Natural Language Understanding (NLU) and Natural Language Generating (NLG). For NLU, we align with the evaluation of LoRA \cite{hu2021lora}, selecting the GLUE Benchmark \cite{wang2018glue} as our evaluation dataset, and compare it with the data reported in \cite{hu2021lora}. For NLG, to validate the effectiveness of SIFT on much larger models, we choose instruction-tuning as our test case. We adopt Llama \cite{touvron2023llama} as our backbone models, use the alpaca dataset \cite{alpaca} for instruction-tuning, and conduct evaluations on benchmarks such as MMLU \cite{hendrycks2020measuring} and HumanEval \cite{chen2021evaluating}. In the experiments, similar to \cite{hu2021lora}, we only consider updating the parameters in the self-attention modules. By adjusting the sparsity rate of SIFT, we ensure a consistent number of trainable parameters. See Appendix \ref{sec_exp_hp} for detailed hyper-parameter settings, categories scores of MMLU and generated samples of HumanEval.
\vspace{0.2cm}\\
In Section \ref{further_analysis}, we conduct further experimental analysis towards SIFT. Since SIFT selects components to update through gradient analysis, we compare SIFT with random component selection in Section \ref{sec_sift_random}. Additionally, as we heuristically choose to perform sparse updates on q,k,v and o, we compare the specific performance of applying SIFT in different parameters. Finally, in Section \ref{sparse_rate}, we explore the impact of sparsity rate on the downstream task, investigating what is the limit of sparsity rate required for an acceptable performance in the downstream task. The evaluation includes the following baselines:
\vskip 0.01in
\textbf{Full fine-tune}: All pre-trained parameters are updated.
\vskip 0.01in
\textbf{Adapter}: \cite{houlsby2019parameter} proposes to adapt pre-trained models to downstream tasks by inserting trainable Adapter modules into the self-attention layers while keeping the pre-trained parameters frozen. These modules have a bottleneck structure with non-linear function. We denote this original version as $\text{Adpt}^H$. Based on \cite{houlsby2019parameter}, \cite{pfeiffer2020adapterfusion} proposed a more efficient version of the Adapter modules, referred to as $\text{Adpt}^P$.
\vskip 0.01in
\textbf{LoRA}\cite{hu2021lora}: freezes the weights of the pre-trained model and adds trainable rank decomposition matrices in parallel to each layer of the Transformer architecture.
\subsection{GLUE Benchmark}
We evaluate SIFT on the GLUE benchmark \cite{wang2018glue}. We select RoBERTa \cite{liu2019roberta} as the backbone model for testing. The general experimental setup is consistent with \cite{hu2021lora} and we apply a 0.8\% sparsity rate with SIFT to the Query, Key, Value, and Output projection in the self-attention module to ensure the number of trainable parameters is consistent with the compared baselines. Table \labelcref{glue} reports the results of pre-trained RoBERTa-Large with different fine-tuning methods on the GLUE benchmark and the number of trainable parameters (excluding the classifier). The results of each run are taken from the best epoch. Table \labelcref{glue} reveals that SIFT performs competitive on the GLUE benchmark compared to full fine-tuning and other PEFT methods.

\subsection{Instruction-tuning}
\citet{wei2021finetuned} proposes to perform supervised fine-tuning on top of pre-trained language models to make them align instructions and generate more meaningful content. \citet{alpaca} fine-tunes pre-trained Llama on a 52k instruction-following dataset (Alpaca), which is generated by the technique introduced by \citet{wang2022self}. We fine-tune Llama with Alpaca in different ways including full fine-tuning, LoRA and our SIFT. We evaluate language models on two benchmarks including MMLU and HumanEval. For MMLU, the evaluation metric is the accuracy of the generating answer in zero-shot setting, while HumanEval is evaluated by pass@1 and pass@10 metric. Higher values are better for all metrics. Table \labelcref{instruction-tuning} demonstrates that sparse structures like SIFT can inject  knowledge into pre-trained LLMs and exhibit a competitive performance.
\begin{table}[h]
\vskip -0.08in
\caption{Performances of fine-tuned Llama on Alpaca dataset with different methods. We present the accuracy on the MMLU task under zero-shot setting, as well as the pass@1 and pass@10 metrics on the HumanEval task. }
\label{instruction-tuning}
\begin{center}
\begin{small}
\begin{sc}
\setlength{\tabcolsep}{4pt}
\begin{tabular}{cccccc}
\toprule
Model & Method & Num & MMLU & HumanEval \\
\midrule
\multirow{4}{*}{Llama-7B} & Vanilla & \diagbox{}{} & 32.7 & 11.3/16.5  \\
          & Full & 6.74B & \textbf{42.0} & 11.1/19.5  \\
          & LoRA & 0.32B & \underline{40.7} & \textbf{11.8}/\underline{20.1} \\
          & SIFT & 0.32B & \underline{40.7} & \underline{11.6}/\textbf{20.7}      \\
\midrule
\multirow{4}{*}{Llama-13B} & Vanilla & \diagbox{}{} & 43.6 & 14.3/24.4 \\
          & Full & 13.0B & \textbf{48.8} & \textbf{15.7}/22.0 \\
          & LoRA & 0.5B & \underline{46.7} & 14.6/25.6 \\
          & SIFT & 0.5B & \underline{46.7} & \underline{15.5}/\textbf{26.2}     \\
          \midrule
\multirow{3}{*}{Llama-33B} & Vanilla & \diagbox{}{} & 54.3 & 20.9/\underline{}{36.0} \\
          & LoRA & 0.98B & \textbf{55.9} &  \underline{21.6}/33.5    \\
          & SIFT & 0.96B & \underline{55.7} &  \textbf{24.1}/\textbf{37.2}     \\
\bottomrule
\end{tabular}
\end{sc}
\end{small}
\end{center}
\vskip -0.2in
\end{table}

\subsection{Further Analysis}\label{further_analysis}
We primarily conduct further experimental analysis towards SIFT involving comparison of performance differences when applying SIFT on different parameters, as well as the differences between SIFT gradient-based component selection and random selection. Finally, we discuss the impact of the sparsity rate on the performance of downstream tasks.

\subsubsection{SIFT Vs. Random}\label{sec_sift_random}
\begin{table}[t]
\caption{Performances of applying SIFT and random selection on different parameters. $W_q, W_k, W_v, W_o$ represent the query, key, value, and output parameters in the attention module  respectively and $W_{All}$ denotes all parameters in the attention module (excluding LayerNorm).}
\label{sift_random}
\begin{center}
\begin{small}
\begin{sc}
\begin{tabular}{clrcccc}
\toprule
Type& Weight & Num & RTE &MRPC &Avg. \\
\midrule
\multirow{5}{*}{SIFT} &$W_V$ & 0.81M & \underline{85.9} & \textbf{90.4} & \textbf{88.0}\\
&$W_{Q,V}$ & 0.81M & 85.2 & \underline{90.2} & 87.7\\
&$W_{Q,K,V}$ & 0.83M & 84.8 & 88.7 & 86.8\\
&$W_{Q,K,V,O}$ & 0.80M & \underline{85.9} & \textbf{90.4} & \textbf{88.0}\\
&$W_{All}$ & 0.80M & \textbf{86.6} &89.0 & \underline{87.8}\\
\midrule
\multirow{5}{*}{Random} &$W_V$ & 0.81M & \underline{83.0} &\underline{89.2} &\underline{86.1} \\
&$W_{Q,V}$ & 0.81M &  82.3 &89.0 &85.7\\
&$W_{Q,K,V}$ & 0.83M & 81.9 &87.5 &84.7\\
&$W_{Q,K,V,O}$ & 0.80M & \textbf{83.8} &\textbf{89.5} &\textbf{86.5}\\
&$W_{All}$ & 0.80M & \underline{83.0} & 88.2 &85.6\\
\bottomrule
\end{tabular}
\end{sc}
\end{small}
\end{center}
\vskip -0.2in
\end{table}

SIFT selects the component with the highest absolute gradient value in the first batch for the subsequent updates. A natural question arises: does this selection have an advantage over random selection? Additionally, in previous experiments, we heuristically choose the $W_q, W_k, W_v$ and $W_o$ in the attention modules for sparse updating. Under the condition of maintaining the same number of trainable parameters, including more modules to apply SIFT means that each module has fewer trainable components. We want to know whether having fewer modules with more components or more modules with fewer components performs better in the fine-tuning process. In Table \labelcref{sift_random}, we compare the performance of SIFT and random selection in updating different modules. We choose two small datasets from the GLUE benchmark, RTE and MRPC, for evaluation. The final performances do not show a consistent conclusion in the choice of different update modules. Therefore, balancing the number of components and modules, we choose to update $W_q, W_k, W_v$ and $W_o$ as our final choice. Compared to random selection, SIFT gradient-based selection method shows a clear advantage in the final performance of downstream tasks. However, although not as effective as SIFT in terms of performances, sparse updates in randomly selected components can still effectively adapt pre-trained models to downstream tasks, with only a small performance gap. This further confirms that due to the shift of the prior induced by pre-training, the model is away from non-expressive submanifolds in parameter space, resulting in redundancy across dimensions. Searching along partial dimensions is sufficient to find a relatively good solution.

\subsubsection{Sparsity Rate Analysis}\label{sparse_rate}

\begin{figure}[h]
\begin{center}
\includegraphics[width=0.32\textwidth]{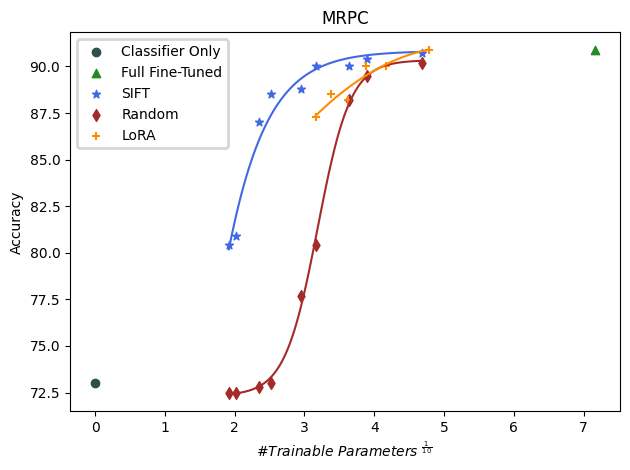}
\caption{Performances of the fine-tuned pre-trained RoBERTa-Large vs different methods and the numbers of trainable parameters. Classifier only indicates that only the classifiers are fine-tuned, and the rest of the methods update the classifiers along with the parts of the pre-trained model.}
\label{fig_sr}
\end{center}
\vskip -0.2in
\end{figure}

We investigate the efficiency of different methods, that is, the trend of performance with the number of trainable parameters (budget). We compare performances of SIFT, Random selection and LoRA with varied budgets on MRPC dataset. Figure \ref{fig_sr} show that SIFT is able to adapt the downstream task more efficiently in a less budget, while the performance gap between SIFT and LoRA is smoothed out as the budget increases. Consistent with the conclusion in Section \ref{sec_sift_random}, we find that when the number of parameters reaches a certain threshold, Random is also able to adapt the downstream task, but it is significantly inferior to SIFT and LoRA in terms of performances in less budgets.

\section{Conclusion}
In this paper, we adopt PAC-Bayesian generalization error bounds to analyze the pre-training-fine-tuning paradigm and attribute the performance differences to the prior distribution shift. We verify the shift of the prior through investigating differences of pre-trained models in loss landscapes and gradient distributions. Pre-trained models, compared to models trained from scratch, exhibit sharper oscillations in the loss landscape and gradients display a quasi-sparse property, where partial components dominates the majority of the gradient norm. Due to these differences, we demonstrate that the search space dimensions required for pre-trained models to reach optimal solutions during fine-tuning are compressed, allowing effective adaptation to downstream tasks by fine-tuning only a small portion of parameters. Based on this, we propose a component-sparse fine-tuning strategy, named SIFT. We validate the effectiveness of using SIFT for fine-tuning on varied tasks including GLUE Benchmark and instruction-tuning for large language models. Furthermore, we have further discussions on SIFT, highlighting its efficiency from the comparison of different component selection methods and the performance trends of different methods as it varies with the number of trainable parameters. Overall, we demonstrates the feasibility of fine-tuning language models entirely in a component-sparse way.







\section*{Impact Statement}

The main motivation of this work has been to design a simple but efficient fine-tuning method with a better understanding of pre-trained models. By uncovering certain important features and properties within pre-trained models, such as the quasi-sparsity of gradients proposed in this paper, we aim to utilize resources more efficiently during the fine-tuning process. This work can be of interest to researchers focusing on efficient fine-tuning and sparsity in deep models. There is no foreseeable negative societal/ethical impact of our work at this time.

\nocite{langley00}

\bibliography{example_paper}
\bibliographystyle{icml2024}


\newpage
\appendix
\onecolumn
\section{A Primer on PAC-Bayesian Generalization Error Bounds.}\label{appendix-pac-bayes}
Due to limited pages of the main paper, all statements regarding PAC-Bayesian theory in the main text are kept as concise as possible. Therefore, in this section, we primarily address readers unfamiliar with PAC Learning or PAC-Bayesian Theorems. We will rigorously restate the theorems mentioned in the main paper, providing proofs for the simpler theorems and referencing related papers for the complex one.
\vspace{0.2cm} \\
The theorems in this section are introduced in previously published works. The presented overview here serves as a reminder and provides references for the readers as needed. The notation in this section remains consistent with that in the main paper.
\begin{theorem}
\label{thm:thm1}
If $\mathcal{H}$ is a finite hypothesis space, for any hypotheses $h \in \mathcal{H}$, any loss functions $l$ bounded in $[0,1]$, $0 < \delta < 1$, with a probability at least $1-\delta$ over the selection of n i.i.d. samples,
$$\hat{\mathcal{R}}(h)\leq\mathcal{R}(h)+\sqrt{\frac{log(|\mathcal{H}|)+log(1/\delta)}{2n}}$$.
\end{theorem}
\begin{proof}
    Using Hoeffding's inequality, for any $\epsilon >0$,
    $$\mathbb{P}\{\Delta(h)=\hat{\mathcal{R}}(h)-\mathcal{R}(h)\geq \epsilon\}\leq e^{-2n\epsilon^2}$$
    Applying the Union Bound, $$\mathbb{P}\{\exists h \in \mathcal{H}, \Delta(h)>\epsilon\}\leq \underset{h\in\mathcal{H}}{\sum}\mathbb{P}\{\Delta(h)>\epsilon\}\leq |\mathcal{H}|e^{-2n\epsilon^2}$$
    Let $|\mathcal{H}|e^{-2n\epsilon^2}=\delta$, and solve the equation for $\epsilon$, we have $\epsilon=\sqrt{\frac{log(|\mathcal{H}|)+log(1/\delta)}{2n}}$.\\
\end{proof}

\begin{theorem}
\label{thm:thm2}
(McAllester, 1998). If $\mathcal{H}$ is a finite hypothesis space, for any probability distribution $\mathcal{P}$ over $\mathcal{H} $ that assigns non-zero values, any hypotheses $h \in \mathcal{H}$, any loss functions $l$ bounded in $[0,1]$, $0 < \delta < 1$, with a probability at least $1-\delta$ over the selection of n i.i.d. samples,
$$\hat{\mathcal{R}}(h)\leq\mathcal{R}(h)+\sqrt{\frac{log(1/\mathcal{P}(h))+log(1/\delta)}{2n}}$$.
\end{theorem}
\begin{proof}
    For any fixed $h\in\mathcal{H}$, considering the Chernoff Bound, we have $$\mathbb{P}\{\Delta(h)>\sqrt{\frac{log(1/\mathcal{P}(h))+log(1/\delta)}{2n}}\}=\mathbb{P}\{\Delta^2(h)>\frac{log(1/\mathcal{P}(h))+log(1/\delta)}{2n}\} \leq \mathcal{P}(h)\delta$$
    Then using the Union Bound,
    $$\mathbb{P}\{\exists h \in \mathcal{H}, \Delta(h)>\sqrt{\frac{log(1/\mathcal{P}(h))+log(1/\delta)}{2n}}\}\leq \underset{h\in\mathcal{H}}{\sum}\mathbb{P}\{\Delta(h)>\sqrt{\frac{log(1/\mathcal{P}(h))+log(1/\delta)}{2n}}\} \leq \underset{h\in\mathcal{H}}{\sum}\mathcal{P}(h)\delta=\delta $$

\end{proof}

\begin{theorem}
\label{thm:thm3}
(McAllester, 2003). For any probability distribution $\mathcal{P}$ on a possibly uncountable hypothesis space $\mathcal{H}$ and any measurable loss function $l$, with a probability at least $1-\delta$,
$$\forall \mathcal{Q}\ on\ \mathcal{H}, \underset{h \sim \mathcal{Q}}{\mathbb{E}}[R(h)] \leq \underset{h \sim \mathcal{Q}}{\mathbb{E}}[\hat{R}(h)] + \sqrt{\frac{\mathbb{KL}(Q||P)+\log({n/\delta})+2}{2n-1}} $$
\end{theorem}

A weak version of Theorem \ref{thm:thm3} is first introduced in \cite{mcallester1999pac}. The complete proof of Theorem \ref{thm:thm3} is in \cite{mcallester2003pac}. \cite{maurer2004note,catoni2007pac,thiemann2017strongly} have developed refinements on it.

\section{Experimental Details}\label{sec_exp_hp}
\subsection{GLUE Benchmark Setting}\label{sec_glue_hp}
The table \ref{hp_glue} shows our hyper-parameters settings on the GLUE benchmark experiments.
\begin{table}[H]
\caption{The hyperparameters of the GLUE Benchmark experiments}
\label{hp_glue}
\begin{center}
\begin{small}
\begin{sc}
\renewcommand{\tabcolsep}{3pt}
\begin{tabular}{cccccccccc}
\toprule
\diagbox{Method}{Dataset}  &  & MNLI & SST-2 & MRPC & CoLA & QNLI & QQP & RTE & STS-B  \\
\midrule
\multirow{3}{*}{} & Optimizer & \multicolumn{8}{c}{AdamW}\\
                    & Warmup Ratio & \multicolumn{8}{c}{0.06}\\
                    & LR Schedule & \multicolumn{8}{c}{Linear}\\
\midrule
\multirow{7}{*}{SIFT} & Batch size & \multicolumn{8}{c}{32}\\
                    & Epochs &10&15&20&20&10&20&20&30\\
                    & Learning rate &7e-5&7e-5&7e-5&7e-5&5e-5&7e-5&7e-5&8e-5 \\
                    & Weight decay & \multicolumn{8}{c}{0.1}\\
                    & Max Seq. length & \multicolumn{8}{c}{512}\\
                    & Sparsity rate & \multicolumn{8}{c}{0.8\%}\\
                    & SIFT modules & \multicolumn{8}{c}{$W_q, W_k, W_v, W_o$}\\

\bottomrule
\end{tabular}
\end{sc}
\end{small}
\end{center}
\end{table}
\subsection{Instruction-tuning Setting}\label{sec_it_hp}
The table \ref{hp_it} shows our hyper-parameters settings on the Instruction-tuning experiments.
\begin{table}[H]
\caption{The hyper-parameters of the Instruction-tuning experiments}
\label{hp_it}
\begin{center}
\begin{small}
\begin{sc}
\renewcommand{\tabcolsep}{3pt}
\begin{tabular}{ccccc}
\toprule
\diagbox{Method}{Model}  &  & Llama-7b & Llama-13b & Llama-33b \\
\midrule
\multirow{3}{*}{} & Optimizer & \multicolumn{3}{c}{AdamW}\\
                    & Warmup Ratio & \multicolumn{3}{c}{0.03}\\
                    & LR Schedule & \multicolumn{3}{c}{Linear}\\
\midrule
\multirow{5}{*}{Full} & Batch size & \multicolumn{2}{c}{128} & \diagbox{}{}\\
                    & Epochs &\multicolumn{2}{c}{3}&\diagbox{}{}\\
                    & Learning rate &\multicolumn{2}{c}{2e-5}&\diagbox{}{}\\
                    & Weight decay & \multicolumn{2}{c}{0.}&\diagbox{}{}\\
                    & Max Seq. length & \multicolumn{2}{c}{2048}&\diagbox{}{}\\

\midrule
\multirow{7}{*}{LoRA} & Batch size & \multicolumn{3}{c}{128}\\
                    & Epochs &\multicolumn{3}{c}{3}\\
                    & Learning rate &2e-4 &2e-4 & 1e-4\\
                    & Weight decay & \multicolumn{3}{c}{0.}\\
                    & Max Seq. length & \multicolumn{3}{c}{2048}\\
                    & LoRA rank & \multicolumn{3}{c}{$r_q=r_v=128$}\\
                    & LoRA alpha & \multicolumn{3}{c}{256}\\
\midrule
\multirow{7}{*}{SIFT} & Batch size & \multicolumn{3}{c}{128}\\
                    & Epochs &\multicolumn{3}{c}{3}\\
                    & Learning rate &2e-4&9e-5 &8e-5\\
                    & Weight decay & \multicolumn{3}{c}{0.}\\
                    & Max Seq. length & \multicolumn{3}{c}{2048}\\
                    & Sparsity rate & 5\% & 4\% &3 \%\\
                    & SIFT modules & \multicolumn{3}{c}{$W_q, W_k, W_v, W_o$}\\

\bottomrule
\end{tabular}
\end{sc}
\end{small}
\end{center}
\end{table}

\subsection{Results of differernt categories in MMLU Benchmark}
\begin{figure}[h]
\begin{center}
\includegraphics[width=0.35\textwidth]{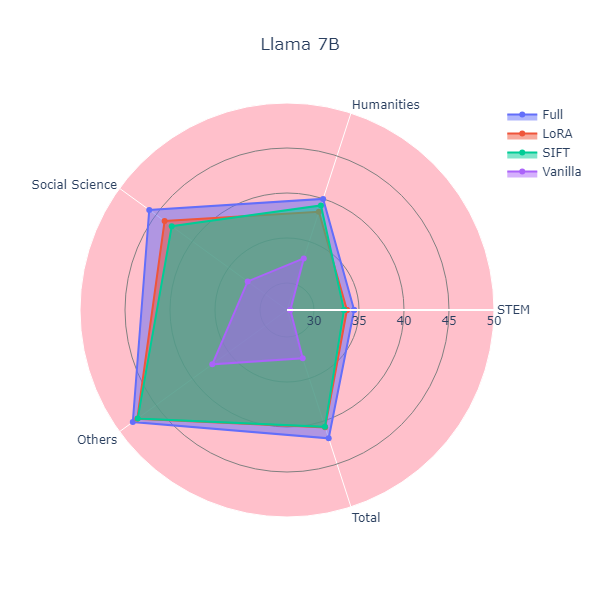}
\caption{MMLU Evaluation results of Llama-7b fine-tuned by different methods}
\label{llama7b}
\end{center}
\end{figure}

\begin{figure}[h]
\begin{center}
\includegraphics[width=0.35\textwidth]{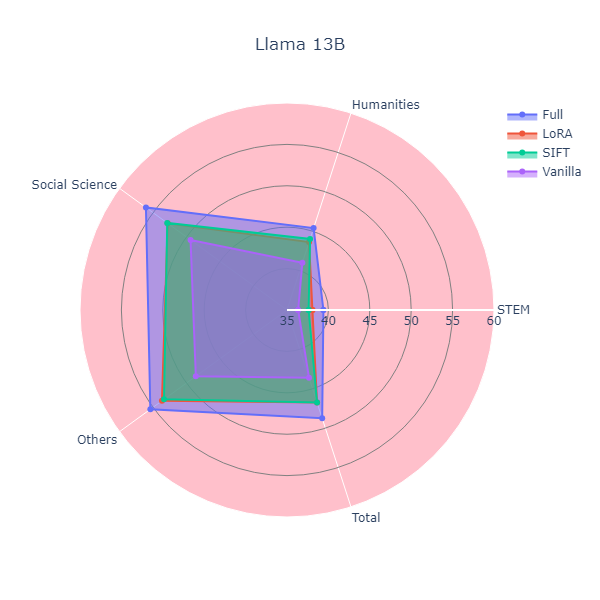}
\caption{MMLU Evaluation results of Llama-13b fine-tuned by different methods}
\label{llama13b}
\end{center}
\end{figure}

\begin{figure}[H]
\vskip 0.2in
\begin{center}
\includegraphics[width=0.35\textwidth]{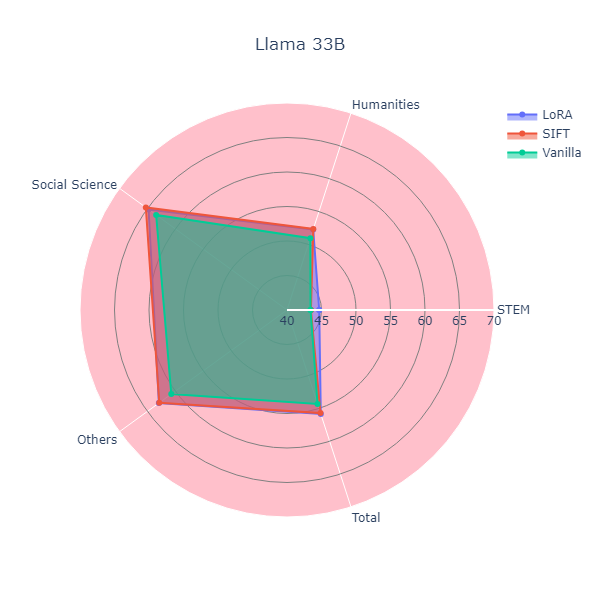}
\caption{MMLU Evaluation results of Llama-33b fine-tuned by different methods}
\label{llama33b}
\end{center}
\vskip -0.2in
\end{figure}

Figures \ref{llama7b}, \ref{llama13b}, and \ref{llama33b} serve as a supplement to Table \ref{instruction-tuning}, illustrating the score comparisons for various categories on MMLU benchmark under different configurations.

\subsection{Generated Code Samples in HumanEval Benchmark}

\begin{table}[H]
\caption{Generated Code Samples.}
\label{code_sample}
\begin{center}
\begin{small}
\begin{sc}
\begin{tabular}{cc>{\centering\arraybackslash}m{0.5\textwidth}c}
\toprule
Model& Method & Generated Samples & Passed\\
\midrule
\multicolumn{2}{c}{Prompt} &
\begin{verbatim}
from typing import List\n\n\n
def below_zero(operations: List[int]) -> bool:\n
\"\"\" You're given a list of deposit and withdrawal
operations on a bank account that starts with\n
zero balance. Your task is to detect if at any point
the balance of account fallls below zero, and\n
at that point function should return True.
Otherwise it should return False.\n
>>> below_zero([1, 2, 3])\n
False\n
>>> below_zero([1, 2, -4, 5])\n
True\n
\"\"\"\n",
"entry_point": "below_zero", "canonical_solution":
"    balance = 0\n\n
for op in operations:\n
balance += op\n
if balance < 0:\n
return True\n\n
return False\n",
"test": "\n\nMETADATA = {\n    'author': 'jt',\n
'dataset': 'test'\n}\n\n\ndef check(candidate):\n
assert candidate([]) == False\n
assert candidate([1, 2, -3, 1, 2, -3]) == False\n
assert candidate([1, 2, -4, 5, 6]) == True\n
assert candidate([1, -1, 2, -2, 5, -5, 4, -4]) == False\n
assert candidate([1, -1, 2, -2, 5, -5, 4, -5]) == True\n
assert candidate([1, -2, 2, -2, 5, -5, 4, -4]) == True\n

\end{verbatim} &\\
\midrule
\multirow{18}{*}{Llama-7B} & Vanilla &
\begin{verbatim}
return operations[0] < 0
\end{verbatim} &False\\
                    & Full & 
\begin{verbatim}
if operations[-1] < 0:\n
return True\n
else:\n
return False
\end{verbatim}& False\\
                    & LoRA & 
\begin{verbatim}
if operations[0] < 0:\n
return True\n
else:\n
return False\n
\end{verbatim}
& False\\
                    & SIFT &
\begin{verbatim}
balance = 0\n
for operation in operations:\n
balance += operation\n
if balance < 0:\n
return True\n
return False
\end{verbatim} &True\\
\midrule
\end{tabular}
\end{sc}
\end{small}
\end{center}
\vskip -0.1in
\end{table}


\end{document}